\documentclass[9pt,conference]{IEEEtran}
\IEEEoverridecommandlockouts

\usepackage{cite}
\usepackage{amsmath,amssymb,amsfonts}
\usepackage{algorithmic}
\usepackage{graphicx}
\usepackage{textcomp}
\usepackage{xcolor}
\usepackage{url}
\usepackage{amsthm}
\usepackage{amssymb}
\usepackage{graphicx}
\usepackage{booktabs}
\usepackage{placeins}
\usepackage{multirow}
\usepackage{tabularx}
\usepackage{epstopdf}
\usepackage{color}
\usepackage{subcaption}
\usepackage{amsfonts}
\usepackage{amsmath}
\usepackage{balance}
\usepackage{enumitem}
\usepackage{xcolor}
\usepackage{wrapfig}
\usepackage{graphics}
\usepackage{todonotes}
\usepackage{makecell}
\usepackage{bbm}
\usepackage{bm}
\def\BibTeX{{\rm B\kern-.05em{\sc i\kern-.025em b}\kern-.08em
    T\kern-.1667em\lower.7ex\hbox{E}\kern-.125emX}}
\begin{document}

\title{STAB: Speech Tokenizer Assessment Benchmark\\
}

\author{\IEEEauthorblockN{Shikhar Vashishth$^*$, Harman Singh$^*$, Shikhar Bharadwaj${}^{*}$\thanks{${}^{*}$Equal Contribution.}, Sriram Ganapathy, Chulayuth Asawaroengchai, \\ Kartik Audhkhasi, Andrew Rosenberg, Ankur Bapna, Bhuvana Ramabhadran \\ 
\texttt{\small \{shikharv,hrman,shikharop,srigana,chulayuth,kaudhkhasi,rosenberg,ankurbpn,bhuv\}@google.com}}  \\
{Google}
}

\maketitle

\begin{abstract}
    Representing speech as discrete tokens provides a framework for transforming speech into a format that closely resembles text, thus enabling the use of speech as an input to the widely successful large language models (LLMs). Currently, while several speech tokenizers have been proposed, there is ambiguity regarding the properties that are desired from a tokenizer for specific downstream tasks and its overall generalizability. Evaluating the performance of tokenizers across different downstream tasks is a computationally intensive effort that poses  challenges for scalability. To circumvent this requirement, we present STAB (Speech Tokenizer Assessment Benchmark), a systematic evaluation framework designed to assess speech tokenizers comprehensively and shed light on their inherent characteristics. This framework provides a deeper understanding of the underlying mechanisms of speech tokenization, thereby offering a valuable resource for expediting the advancement of future tokenizer models and enabling comparative analysis using a standardized benchmark.
    We evaluate the STAB metrics and correlate this with downstream task performance across a range of speech tasks and tokenizer choices.
\end{abstract}


\newcommand{\red}[1]{\textcolor{red}{#1}}
\newcommand{\audiopalm}{AudioPaLM}

\newcommand{\refalg}[1]{Algorithm \ref{#1}}
\newcommand{\refeqn}[1]{Equation \ref{#1}}
\newcommand{\reffig}[1]{Figure \ref{#1}}
\newcommand{\reftbl}[1]{Table \ref{#1}}
\newcommand{\refsec}[1]{Section \ref{#1}}

\newcommand{\reminder}[1]{\textcolor{red}{[[ #1 ]]}\typeout{#1}}
\newcommand{\reminderR}[1]{\textcolor{gray}{[[ #1 ]]}\typeout{#1}}

\newcommand{\add}[1]{\textcolor{red}{#1}\typeout{#1}}
\newcommand{\remove}[1]{\sout{#1}\typeout{#1}}

\newcommand{\method}{STAB}

\newcommand{\problem}{DD}
\newcommand{\problemfull}{Document Dating}

\newcommand{\mc}[1]{\mathcal{#1}}
\newcommand{\bmm}[1]{\bm{\mathcal{#1}}}
\newcommand{\real}[1]{\mathbb{R}^{#1}}

\newcommand{\tensor}{\mathcal{X}}
\newcommand{\Real}{\mathbb{R}}

\newcommand{\tuples}{\mathbb{T}}

\newcommand{\argmax}{arg\,max}

\newcommand\norm[1]{\left\lVert#1\right\rVert}

\newcommand{\note}[1]{\textcolor{blue}{#1}}

\newcommand*{\Scale}[2][4]{\scalebox{#1}{$#2$}}%
\newcommand*{\Resize}[2]{\resizebox{#1}{!}{$#2$}}%
\definecolor{officegreen}{rgb}{0.0, 0.5, 0.0}
\def\mat#1{\mbox{\bf #1}}
\section{Introduction}
\label{sec:introduction}
Speech representation learning, the task of developing models that  extract succinct feature representations of speech for downstream tasks, has been area of active interest in the recent years.
Motivated by zero-resource speech processing to develop methods that can learn sub-word or word units directly from unlabeled raw speech~\cite{versteegh2016zero}, several unsupervised methods have been proposed for learning continuous representations~\cite{renshaw2015comparison,hsu2017unsupervised} and discrete acoustic units~\cite{vinyals2015show}. 
Techniques based on predictive coding \cite{chung2020generative} and self-supervision learning such as the class of wav2vec models \cite{wave2vec2}, have been developed to derive quantized representations of audio.  More recently, iterative learning of discrete units and acoustic representations such as HuBERT \cite{hubert} and  joint learning of denoising and self-supervision in wavLM \cite{chen2022wavlm} have shown promising results.

Discrete representations are a natural fit for speech and language given their ability to be represented as a sequence of symbolic, phonetic, graphemic or sub-word/word units. The approach of representing speech in the form of discrete tokens offers a significant advantage by converting speech into a format that mirrors text, thereby leveraging the application of speech as an input for various large language models (LLMs) \cite{llama,palm2}. Furthermore, speech tokens have the ability to capture non-verbal cues such as emotion and rhythm, which contain additional information compared to their textual counterparts \cite{speech_token_with_non_verbal1,speech_token_with_non_verbal2}. 
Utilizing discrete speech tokens has proven advantageous in tasks such as automatic speech translation and speech-to-speech translation, while demonstrating comparable performance on automatic speech recognition \cite{audiolm,audiopalm}. This also contributes to the advancement of multimodal LMs \cite{gemini}.

Speech tokenizers optimized for specific downstream task(s) exist~\cite{eloff2019unsupervised}, however, measuring their generalization ability remains a challenging problem.
Assessing the performance of all tokenizers across various downstream tasks is a computationally expensive endeavor that presents challenges for scalability. Additionally, speech tokenizers are often utilized as a black box, with limited examination~\cite{irasl2018} of the nature of the tokens they generate or their adherence to specific properties.
Therefore, it is timely to create a low-compute evaluation benchmark for assessing tokenizers across multiple dimensions. 
Our contributions can be summarized as follows: 

\begin{itemize}[itemsep=2pt,parsep=1pt,partopsep=1pt,leftmargin=*,topsep=1pt]
    \item We propose STAB, a speech tokenizer assessment benchmark which evaluates  capabilities of a given speech tokenizer. 
    \item STAB presents a cost-effective evaluation approach and holds potential for expediting research on speech tokenization.
    \item Through extensive experiments, we demonstrate that STAB provides a reliable indication of the speech tokenizer's performance on a range of  downstream tasks.
\end{itemize}

\section{Related Work}
\label{sec:related_works}

\begin{figure*}[t]
	\centering
	\includegraphics[width=0.9\textwidth]{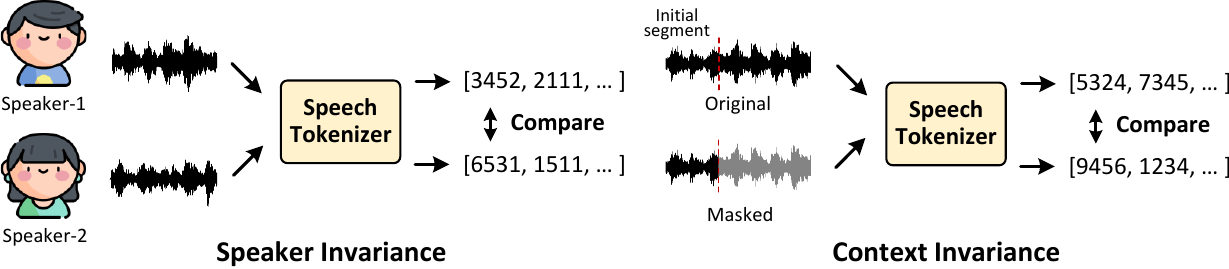}
	\caption{\label{fig:overview}\small STAB's Invariance Dimensions: (left) illustrates speaker invariance, comparing the tokenization of the same sentence spoken by two different speakers. (right) demonstrates context invariance, comparing the tokenization of an initial segment of the speech signal with and without the availability of the original context. Refer to Section \ref{sec:details} for more details.}
	\vspace{-0.15in}
\end{figure*}

\textbf{Speech Tokenization:}
Self-supervised learning for speech historically relied on contrastive loss on audio embeddings, as exemplified by wav2vec \cite{wave2vec}.
Vq-wav2vec \cite{vq-wav2vec} and DiscreteBERT \cite{discrete_bert} introduced tokenization based objectives for learning better speech representations.
Following this, HuBERT \cite{hubert} introduced iterative refinement of speech tokens within a masked language modeling (MLM) framework.
W2v-BERT \cite{w2v_bert} combined the benefits of the contrastive approaches and MLM with speech tokens in a single model.
Interestingly, BEST-RQ \cite{best_rq} utilizes random projection to generate target tokens.
Hence, speech tokens have become central to self-supervised pre-training models and are typically obtained through methods such as K-means or vector quantization \cite{vector_quantization}.
AudioLM \cite{audiolm} and AudioPaLM \cite{audiopalm} auto-regressively model the speech token sequences derived from clustering representations generated by an audio encoding model. In this study, we assess various speech tokenizers employed in existing methods.

\textbf{Speech Benchmarks}
With the development of various representation learning frameworks, there have also been efforts to evaluate and benchmark speech representations. In the latest edition of the Zero Resource Speech Challenge, evaluations focused on exploring text-less speech language modeling tasks \cite{dunbar2022self}. The speech processing universal performance benchmark (SUPERB) considers a multitude of downstream evaluation tasks that included semantic and para-linguistic tasks \cite{yang2021superb}. An extension to multi-lingual tasks is benchmarked in ML-SUPERB \cite{shi2023ml}. For multitask evaluation in a zero-shot setting, Dynamic-SUPERB~\cite{huang2023dynamic} has been introduced recently. A non-semantic evaluation benchmark, NOSS has also been proposed for audio representations~\cite{shor2020towards}. 

\section{STAB Details}
\label{sec:details}


\subsection{Invariance}
For tasks such as ASR, extracting semantic meaning from speech is crucial. Previous studies have introduced the concept of \textit{semantic} and \textit{acoustic} speech tokenization \cite{audiolm,audiopalm}. Semantic tokens focus solely on extracting semantic information from the speech signal, while acoustic tokens capture other properties such as speaker information, language, and emotion.
Here, we assess the ability of a speech tokenizer to accurately capture semantics by evaluating it along the following dimensions.

\begin{itemize}[itemsep=2pt,parsep=1pt,partopsep=1pt,leftmargin=*,topsep=1pt]
    \item \textbf{Speaker invariance:} Examines variance in tokenization of identical sentences uttered by two different speakers such as comparing a sentence spoken by a female/male speaker.
    


    \item \textbf{Context invariance:} Analyses how tokens are altered when a part of the speech context is masked. This measurement reflects the influence neighborhood of a token.    We compare the tokens extracted from a segment (initial 4 seconds) of the utterance against the same segment with its original context.
    \item \textbf{Language invariance:} Measures the variation in tokenization of the same concept spoken in two different languages. For example, comparing the tokens of \textit{"Cat is drinking the milk"} in English and \textit{"Eine Katze trinkt Milch"} in German.
\end{itemize}

\subsection{Robustness}

We evaluate the resilience of a speech tokenizer to different types of noise and acoustic variations in speech signals. This is crucial for effectively handling real-world data, which may include recordings from a variety of microphones and speakers.

\begin{itemize}[itemsep=2pt,parsep=1pt,partopsep=1pt,leftmargin=*,topsep=1pt]
    \item \textbf{Pitch Change:} Pitch change is a common phenomenon in speech, often resulting from factors such as equipment imperfections or signal processing \cite{pitch_change}. We investigate how tokenization varies when the pitch of a speech signal is modified, while ensuring that the audio remains intelligible.

    \item \textbf{Playback speed:} Modifying the playback speed of audio involves adjusting the rate at which the audio is played. We examine how tokenization changes  when we alter this.

    \item \textbf{Background Noise:} Here, we introduce background noise $(\mathcal{N}(0, v)$ where $v$ is s.t. SNR = 10dB$)$ into the original speech signal and assess the behavior of a speech tokenizer in response to the added noise.

\end{itemize}

\begin{table*}[t!]
\centering
	\small
	\begin{tabular}{clccccccc}
	 \toprule
	&  &  & \multicolumn{3}{c}{\textbf{8k-Tokenizers}} & \multicolumn{3}{c}{\textbf{32k-Tokenizers}} \\    
	\cmidrule(r){4-6} \cmidrule(r){7-9}
    & Dimensions  & Metrics & w2v2 & w2v-BERT & BEST-RQ & USM-v1 & USM-v2 & USM-v3 \\
    \midrule
    \multirow{3}{*}{\textbf{Invariance}} & Speaker Invariance & chrF & 7.5  & 13.0 & 4.7  & 15.8 & {36.6} & 13.9 \\
    & Context Invariance & chrF & 27.8 & 45.6 & 48.4 & {73.2} & 45.8 & {50.9} \\
    & Language Invariance & chrF & 5.2 &  7.5 &  4.6 &  6.9 &  {4.3}  & 4.6 \\
    \midrule
    \multirow{3}{*}{\textbf{Robustness}} & Pitch Change & chrF & 10.2 & 21.4 & 8.1 & 25.1 & {33.3} & 19.7 \\
    & Gaussian Noise (10 dB) & chrF & 42.5 & 48.3 & 39.7 & 67.2 & {72.5} & 57.0 \\
    & Speed Change ($\times$0.8) & chrF & 26.0 & 23.8 & 24.9 & 30.3 & {30.9} & 27.0 \\
    \midrule
    \multirow{3}{*}{\textbf{Compressibility}}  & Huffman Efficiency & \% & 13.9 & 16.9 & 11.4 & 13.5 & 16.8 & 16.0 \\
    & Byte-pair Efficiency & \% & 2.6 & 9.8 & 6.1 & 6.3 & 8.8 & 4.2 \\
    & De-duplication Efficiency  & \% & 1.4 & 9.1 & 10.9 & 4.3 & 6.0 & 4.8 \\
    \midrule
    \multirow{3}{*}{\textbf{Vocabulary}} & Per-language Utilization  & \% & 75.3 & 56.8 & 87.2 & 21.7 & 42.4 & 44.5 \\
    & Overall Utilization  & \% & 99.7 & 95.4 & 99.8 & 47.5 & 99.3 & 96.4 \\
    & Vocabulary Entropy  & Score & 95.0 & 91.0 & 97.3 & 82.8 & 94.4 & 90.6 \\
    \bottomrule
	\end{tabular}
\caption{\label{tbl:stab_main} STAB metrics for several existing speech tokenizers on FLEURS dataset.}
\vspace{-0.15in}

\end{table*}

\subsection{Compressibility}
In natural language processing (NLP), models based on words or subwords have been shown to outperform character-based models  \cite{nlp_subword_better1,nlp_subword_better2}. However, most speech tokenizers tokenize at a level lower than phonemes. Previous studies \cite{speech_subword_better} have shown that training a sentence piece tokenizer on speech sequences yields subword-level tokens, resulting in improvements in downstream tasks. Nevertheless, the degree of compressibility varies among different tokenizers. Hence, we propose the following dimensions to measure this property.

\begin{itemize}[itemsep=2pt,parsep=1pt,partopsep=1pt,leftmargin=*,topsep=1pt]
    \item \textbf{Huffman Encoding Efficiency:} Huffman coding algorithm \cite{huffman} is widely used for lossless data compression. We use the Huffman coding algorithm to compress a corpus of speech sequences of a particular language, following which we calculate the compression efficiency.

    \item \textbf{Byte-pair Encoding Efficiency:} Byte-pair encoding (BPE) \cite{bpe_original, bpe} is a tokenization technique that involves iteratively merging the most frequent pair of consecutive tokens to create new tokens. This merging process is repeated until a predefined vocabulary size is reached. Using BPE, it is possible to learn subword-level tokens by merging repeated patterns found in speech token sequences.
    
    \item \textbf{De-duplication Efficiency:} We assess the compressibility of speech sequences by merging adjacent repeating tokens. 
\end{itemize}

\subsection{Vocabulary}
Here, we evaluate how a speech tokenizer utilizes its vocabulary and how this utilization varies across languages. A larger vocabulary size in a speech tokenizer increases the number of parameters in the Speech Language Models (SLMs). Therefore, it is crucial to analyze how the vocabulary is being used and to ensure that there are no   mode collapse issues. To achieve this, we analyze tokenizers along the following axes,

\begin{itemize}[itemsep=2pt,parsep=1pt,partopsep=1pt,leftmargin=*,topsep=1pt]
    \item \textbf{Per-language Utilization}: We examine the proportion of the total vocabulary utilized for each language, considering a fixed number (500k) of observed tokens.

    \item \textbf{Overall Utilization and Entropy:} We explore the vocabulary utilization across all languages and compute entropy of the vocabulary distribution to evaluate any bias towards a subset of tokens.

    \item \textbf{Vocabulary Distribution Comparison Across Language:} We investigate whether the tokenizer captures relationships among languages, with the hypothesis that a tokenizer designed to consider language similarity should exhibit similar vocabulary distributions for related languages.

\end{itemize}
\section{Experimental Setup}
\label{sec:experiments}

\vspace{-1mm}
\subsection{Datasets}
\label{sec:datasets}
\vspace{-1mm}

\textbf{\method{} Datasets:} 
In our proposed benchmark, we employ the FLEURS dataset \cite{dataset_fleurs}, which is the speech counterpart of the FLoRes-101 machine translation dataset \cite{dataset_flores}. FLEURS comprises 2,000 n-way parallel sentences spoken in 102 languages, enabling evaluation on metrics such as language awareness.
Additionally, we employ the TIMIT dataset \cite{dataset_timit}, which includes recordings of 630 speakers reciting 10 sentences each, accompanied by transcripts for each spoken sentence. This enables us to assess speaker-awareness.

\textbf{Pre-training Datasets:}
In our experiments, we employ the AudioPaLM model \cite{audiopalm} which involves initializing with a pre-trained text decoder (PaLM-2 \cite{palm2}) and subsequently making it multimodal by expanding its vocabulary and training it on a speech-text data mixture. Our data mixture consists of a blend of 75\% original text data \cite{palm2} and 25\% automatic speech recognition (ASR) data sourced from the Babel \cite{dataset_babel}, VoxPopuli ASR, Multilingual Librispeech \cite{dataset_mls}, FLEURS, and YouTube ASR datasets \cite{google_usm}. In total, the speech dataset comprises 221k hours of ASR data spanning across $\sim$100 languages.

\textbf{Evaluation Datasets:}
Along with \method{} benchmark, we evaluate our models on several downstream tasks as well such as ASR, emotion recognition, speaker identification, and intent classification. For ASR, we utilize transcribed VoxPopuli dataset which spans across 14 languages and CoVoST-2 \cite{dataset_covost} dataset for AST. We use IEMOCAP dataset \cite{dataset_iemocap} for emotion recognition and VoxCeleb \cite{dataset_voxceleb} for speaker identification. Since AudioPaLM is a decoder-only model we approach the classification task as a seq2seq task. For all these datasets, we fine-tune our model on their training split followed by evaluation on the corresponding dev/test split.

\vspace{-0.1in}
\subsection{Baseline systems}
\label{sec:baselines}
\vspace{-0.1in}
In our experimental analysis, we compare several speech tokenizers commonly utilized within the research community.
\begin{itemize}[itemsep=2pt,parsep=1pt,partopsep=1pt,leftmargin=*,topsep=1pt]
    \item \textbf{w2v2}: Similiar to Rubenstein et al. \cite{audiopalm}, we employ wav2vec 2.0 \cite{wave2vec2}, which is trained on multilingual data, for encoding speech. Subsequently, a k-means (with k = 8k) is trained on the embeddings generated by the model, and the centroid indices are extracted as semantic tokens.
    \item \textbf{w2v-BERT}: Same as w2v2 with speech encoder replaced by a pre-trained w2v-BERT encoder \cite{w2v_bert} trained using Masked Language Modeling (MLM) objective.
    \item \textbf{BEST-RQ}: Here, we employ MLM-based BEST-RQ model \cite{best_rq} as the speech encoder.
    \item \textbf{USM-v1}: Following Rubenstein et al. \cite{audiopalm}, we employ Google Universal Speech model (USM) \cite{google_usm}, which is trained using MLM objective for encoding speech. For USM-v1 and subsequent tokenizers, the vocabulary size is 32k. 
    
    \item \textbf{USM-v2}  \cite{audiopalm}: Similar to USM-v1, this involves USM but with the inclusion of an auxiliary ASR loss during training. Moreover, instead of K-means, vector quantization \cite{vector_quantization} is used for discretizing representations.
    \item \textbf{USM-v3}: This is identical to USM-v2 tokenizer. However, it utilizes USM trained with spectrogram reconstruction \cite{spectrogram_recon} loss in addition to ASR. 
\end{itemize}
\noindent \textbf{Implementation details:}
\label{sec:hyperparams}
Most of the hyper-parameters are directly adopted from AudioPaLM  \cite{audiopalm}. We report results with models of size 1B, initialized with PaLM-2 checkpoint and pre-trained for 30k steps on our speech-text mixture. For each downstream evaluation task, we fine-tune the pre-trained model on its corresponding training split before evaluation. Please note that no fine-tuning is necessary for \method{}, as metrics can be directly computed over the raw tokens. 

\begin{figure}[t!]
    \centering
    \includegraphics[width=\columnwidth]{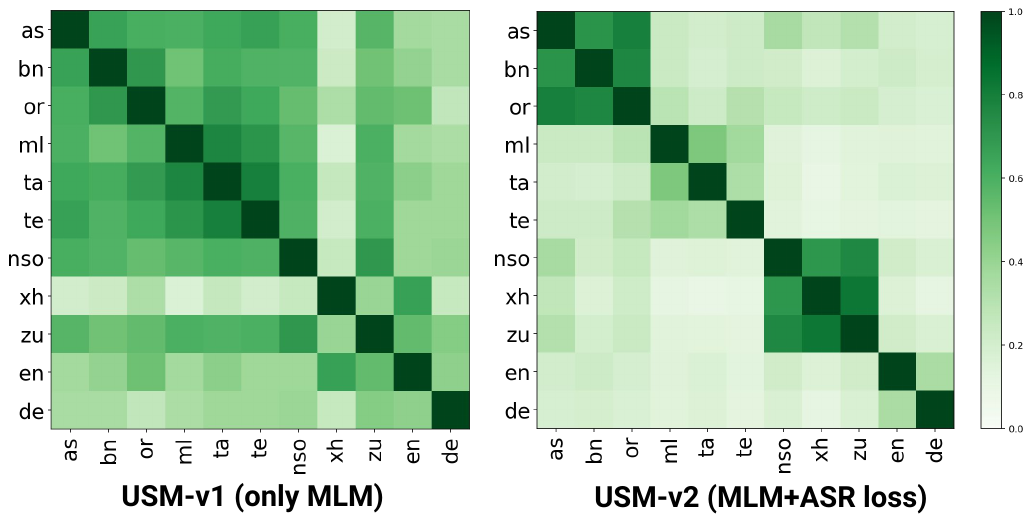}
    \caption{Vocabulary distribution for USM-v1 and USM-v2 tokenizers. The inclusion of ASR loss allows the tokenizer to capture language relatedness. Refer to Section \ref{sec:stab_comparison} for details.}
    \label{fig:vocab_distribution}
    \vspace{-0.1in}
\end{figure}

\section{Results}
\label{sec:results}

\subsection{STAB Performance Comparison}
\label{sec:stab_comparison}
In this section, we evaluate different tokenizers, as outlined in Section \ref{sec:baselines}, on various STAB dimensions. The summary of the results is presented in Table \ref{tbl:stab_main}. As previously described, w2v2 is trained using contrastive loss whereas w2v-BERT, BEST-RQ and USM-v1 are trained using Masked Language Modeling (MLM) loss. Further, USM-v2 incorporates both MLM and Automatic Speech Recognition (ASR) losses and USM-v3 additionally includes reconstruction loss. Please note that the vocabulary size of w2v2, w2v-BERT, and BEST-RQ tokenizers is 8k, whereas USM-based tokenizers utilize a vocabulary of 32k. 
Thus, the majority of our conclusions are drawn from comparisons within the group of tokenizers having the same vocabulary size.

\begin{figure}[t!]
    \centering
    \includegraphics[width=0.95\columnwidth]{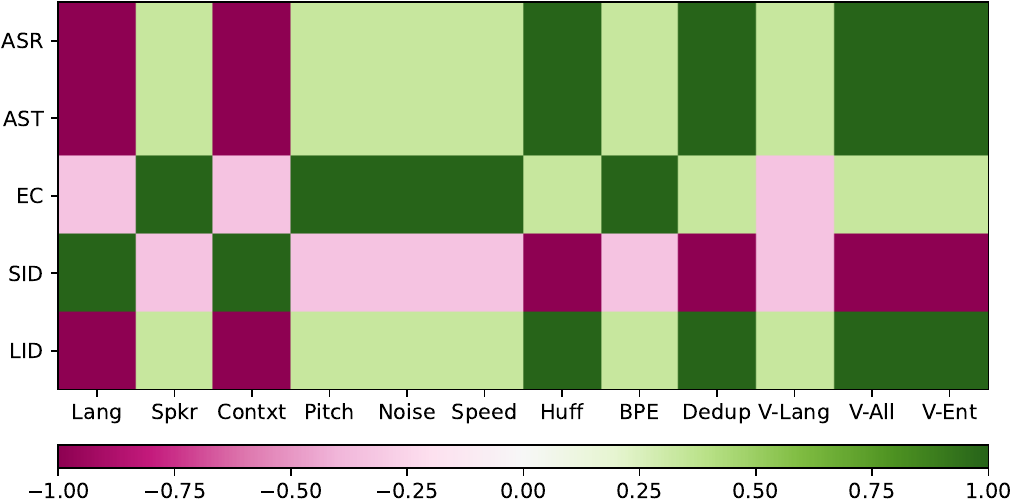}
    \caption{Correlation plot showing the relationship between the STAB metrics (Table~\ref{tbl:stab_main}) and the downstream task performance (Table \ref{tbl:downstream_tasks}). Here, pairs of tokenizers are considered and the correlation is computed between the relative improvements in STAB metrics w.r.t. the relative improvements in task performance.}
    \label{fig:correlation_plot}
    \vspace{-0.2in}
\end{figure}

\textbf{Invariance:} The results demonstrate that the inclusion of ASR loss (such as in USM-v2) makes tokenizers more invariant to speaker information. Moreover, it boosts contextual dependence, as it necessitates a semantic understanding of all frames collectively. This is evident through fall in context invariance metric on USM-v2 among 32k-tokenizers. Further, contrastive loss of w2v2 drastically increases the dependence on context compared to MLM-based losses in w2v-BERT and BEST-RQ. Regarding language invariance, most tokenizers generate distinct token sequences for different languages. However, ASR loss appears to reduce language invariance, as it necessitates generating text in the correct script based on the specific language used in the speech.


\textbf{Robustness:} We observe that USM-based tokenizers exhibit greater robustness to noise compared to other tokenizers, likely due to the more extensive data used during pre-training. Additionally, incorporating ASR loss during training enhances the tokenizers' resilience to noisy speech signals. In contrast, training with a spectrogram reconstruction loss appears to increase the model's susceptibility to noise. Among the 8k-tokenizers, the w2v-BERT tokenizer demonstrates superior noise robustness relative to its counterparts.

\textbf{Compressibility:} The results indicate that 8k-tokenizers demonstrate higher compressibility compared to 32k-tokenizers, which can be attributed to their smaller vocabulary size. Among 8k-tokenizers, w2v-BERT exhibits higher overall compressibility. Additionally, similar to previous findings, incorporating ASR loss enhances tokenizer compressibility.

\textbf{Vocabulary:}
Among all 32k-tokenizers, USM-v1 exhibits the lowest per-language and overall vocabulary utilization. This is attributed to its use of K-means quantization, in contrast to the vector quantization employed by USM-v2 and USM-v3. This indicates that simple K-means representation is ineffective in fully utilizing the entire vocabulary, potentially resulting in the wastage of model parameters. The vocabulary utilization among 8k-tokenizers is higher given their smaller vocabulary. 

\textbf{Language Relationships:} The ASR loss enhances tokenizer's awareness of language relationships. As shown in Figure \ref{fig:vocab_distribution}, USM-v2 exhibits a higher similarity in vocabulary distribution across closely related languages, a characteristic not elicited by tokens from the USM-v1 tokenizer. This demonstrates the potential of ASR-trained tokenizers to exhibit higher levels of cross-lingual knowledge transfer.

\vspace{-0.05in}
\subsection{Correlation with Downstream tasks}
\label{sec:downstream_eval}
\vspace{-0.05in}
We evaluate various tokenizers on multiple downstream tasks: Automatic Speech Recognition (ASR), Automatic Speech Translation (AST), Emotion Classification (EC), Speaker Identification (SID), and Language Identification (LID). For each task, we fine-tune our already pre-trained models on the training split of corresponding dataset before evaluation. 

The results on downstream tasks are summarized in Table \ref{tbl:downstream_tasks}. Overall, we find that STAB metrics correlates well with the performance on downstream tasks. On ASR and AST tasks, w2vBERT and USM-v2, which are more speaker invariant and robust to noise, perform best in their categories. On the contrary, the tokenizers which have lower speaker invariance performs better on speaker identification tasks as expected. 
Previous studies \cite{emotion_asr} on emotion classification using IEMOCAP dataset have shown that utilizing the output of an ASR system yields better results compared to models that directly use the speech modality. Our findings support this observation, as w2v-BERT and USM-v2 outperform other tokenizers in our experiments. USM-v2 also captures language similarity better, as shown in Figure \ref{fig:vocab_distribution}, which reflects in its improved language identification performance. 

For identifying the coupled relationship between the STAB metrics (Table~\ref{tbl:stab_main}) and the downstream tasks   (Table~\ref{tbl:downstream_tasks}), we consider pairs of tokenizers (eg. USM-v1, USM-v2). For this pair, we compute correlation between the binarized relative improvements in a STAB metric and the relative improvements in a downstream task performance. In this manner, the correlation plot is generated (Figure~\ref{fig:correlation_plot}) using average correlation over all 32k-tokenizer pairs for different choices of STAB metrics and downstream tasks. As seen here, the ASR and AST tasks follow an identical trend with vocabulary utilization metrics showing the maximal correlation while language/context invariance is seen to have the maximal negative correlation. The LID task also shows a similar trend. The EC task shows the highest correlation for speaker and noise invariance, which essentially allows the model to focus on emotion related cues in the tokenized audio signal. The SID task shows somewhat of an opposite trend to most of the other tasks considered, where the language and  context invariance are positively correlated while the overall vocabulary utilization is negatively correlated with the SID performance.  
These findings illustrate that STAB metrics correlate with downstream tasks and   offers insights into a tokenizer's performance on downstream applications.

\begin{table}[t!]
\centering
	\begin{tabular}{lccccc}
	    \toprule
        Tokenizers & ASR  & AST & EC & SID & LID \\
        \cmidrule(r){2-2} \cmidrule(r){3-3} \cmidrule(r){4-6}
         & WER~$\downarrow$  & BLEU~$\uparrow$ & \multicolumn{3}{c}{Accuracy $\uparrow$} \\
        \midrule
        w2v2 & 73.8 & 2.4 & 50.8 & 65.0 & 16.0 \\ 
        w2v-BERT & 53.4 & 7.4 & 55.0 & 38.2 & 28.8 \\ 
        BEST-RQ & 66.6 & 4.1 & 54.0 & 49.8 & 17.2 \\ 
        USM-v1 & 49.3 & 3.6 & 55.9  & 53.0  & 79.4 \\ 
        USM-v2 & 11.8 & 16.8 & 60.0 & 16.3 & 97.1 \\ 
        USM-v3 & 16.5 & 10.2 & 50.9 & 24.6 & 91.8 \\ 
		\bottomrule
    \end{tabular}

\caption{\label{tbl:downstream_tasks}Evaluation on downstream tasks: Speech Recognition (ASR), Speech Translation (AST), Emotion Classification (EC), Speaker (SID), and Language Identification (LID).}
\vspace{-0.2in}
\end{table}

\noindent \textbf{Cost-Effectiveness of STAB:}
For any tokenizer, each STAB metric requires less than 15 minutes of CPU compute on our Apache Beam based implementation. In contrast, evaluating each tokenizer for a downstream task involves approximately 16 hours of pre-training on 256 accelerated hardware chips across multiple datasets, followed by 22 hours of fine-tuning on 128 accelerated hardware chips. Hence, STAB is at least 100x more efficient in terms of compute and data resources compared to downstream evaluation. Consequently, the proposed benchmark has the potential to be a valuable tool in advancing the design of speech tokenizers.

\vspace{-0.05in}
\section{Conclusion}
\label{sec:conclusion}
In this paper, we introduced STAB (Speech Tokenizer Assessment Benchmark), a comprehensive benchmark for evaluating speech tokenizers and illuminating their inherent characteristics. The benchmark offers a deeper understanding of the inner workings of a speech tokenizer, and STAB metrics correlate with the performance on several downstream tasks. STAB is 100x more efficient in terms of compute and data than using downstream tasks to compare speech tokenizers, making it a potential catalyst for the development of speech tokenizers.

\bibliographystyle{IEEEtran}
\bibliography{mybib}

\end{document}